\documentclass{article}

 \usepackage[preprint,nonatbib]{nips_2018}

\usepackage[utf8]{inputenc} 
\usepackage[T1]{fontenc}    
\usepackage{hyperref}       
\usepackage{url}            
\usepackage{booktabs}       
\usepackage{amsfonts}       
\usepackage{nicefrac}       
\usepackage{microtype}      
\usepackage{graphicx}
\usepackage{amsmath,amssymb} 
\usepackage{color}
\usepackage{subfig}
\usepackage{algorithm}
\usepackage{algpseudocode}

\title{Zero-Shot Transfer VQA Dataset}

\author{
  Yuanpeng Li, Yi Yang, Jianyu Wang \\
  \texttt{\{liyuanpeng01,yangyi05,wangjianyu02\}@baidu.com}\\
  Baidu Research, USA
  \And
  Wei Xu\\
  \texttt{wei.xu@horizon.ai}\\
  Horizon Robotics Research, USA
}

\begin{document}

\maketitle

\begin{abstract}
Acquiring a large vocabulary is an important aspect of human intelligence.
One common approach for human to populating vocabulary is to learn words during reading or listening, and then use them in writing or speaking. This ability to transfer from input to output is natural for human, but it is difficult for machines.
Human spontaneously performs this knowledge transfer in complicated multimodal tasks, such as Visual Question Answering (VQA).
In order to approach human-level Artificial Intelligence, we hope to equip machines with such ability.
Therefore, to accelerate this research, we propose a new {\em zero-shot transfer VQA} (ZST-VQA) dataset by reorganizing the existing VQA v1.0 dataset in the way that during training, some words appear only in one module (i.e. questions) but not in the other (i.e. answers). In this setting, an intelligent model should understand and learn the concepts from one module (i.e. questions), and at test time, transfer them to the other (i.e. predict the concepts as answers).
We conduct evaluation on this new dataset using three existing state-of-the-art VQA neural models. Experimental results show a significant drop in performance on this dataset, indicating existing methods do not address the zero-shot transfer problem.
Besides, our analysis finds that this may be caused by the implicit bias learned during training.

\end{abstract}

\section{Introduction}
With the fast development of deep learning~\cite{krizhevsky2012imagenet,simonyan2014very,he2016deep}, Artificial Intelligence achieved human level in many domains~\cite{silver2017mastering,OpenAI_dota}. However, current AIs are only designed for specific tasks, and still far from human's general intelligence. A way to make a human-like machine is to make them learn as the human does so that it is important to understand how the human learns. One characteristic of human learning is the ability to transfer from input to output. For example, to populate vocabulary, people learn words during reading and listening, and use them in writing and speaking. This transferability from input to output is natural for human, but it is difficult for machines.

\begin{figure}[!t]
  \centering
  \subfloat[Example of zero-shot answer task (ZSA). A zero-shot word ``sunglasses'' appears in question but not in answer during train, and it appears in answer for test.]{\includegraphics[width=0.49\textwidth]{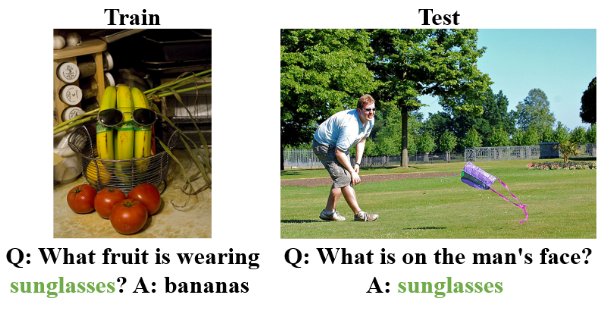}\label{fig:zsa_example}}
  \hfill
  \subfloat[Example of zero-shot question task (ZSQ). A zero-shot word ``chair'' appears in answer but not in question during train, and it appears in question for test.]{\includegraphics[width=0.49\textwidth]{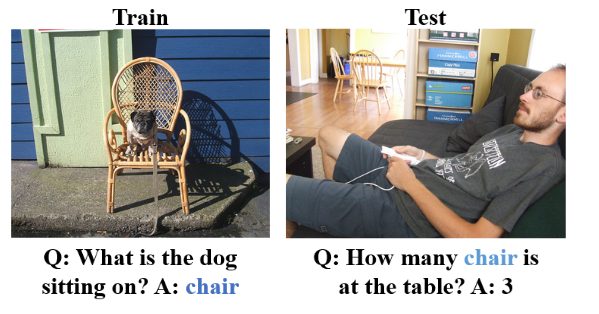}\label{fig:zsq_example}}
  \caption{Examples of ZSA and ZSQ tasks. Transferring learned words from questions to answers or from answers to questions is required in these tasks.}
\end{figure}

Human spontaneously performs this transfer with language compositionality~\cite{gelder1990compositionality} in complicated multimodal tasks, such as VQA. 
For example, in Fig.~\ref{fig:zsa_example}, in order to learn from the question-answer pair [``What fruit is wearing sunglasses?'', ``bananas''], one needs to understand the concept of ``sunglasses''. Then for another question [``What is on the man's face?''], humans can provide the correct answer ``sunglasses'' even though they have never seen it as an answer during training. 
Similarly, humans can also transfer concepts learned from answers to questions.
Although VQA achieved high performances in standard benchmarks~\cite{antol2015vqa,gao2015you,johnson2017clevr,kafle2017visual,gupta2017survey},
little investigation has been made to address abilities for zero-shot transfer learning.

To facilitate this study, we create a new dataset, named as {\em zero-shot transfer VQA} (ZST-VQA), by rearranging the original VQA v1.0 dataset~\cite{antol2015vqa}. The dataset includes two tasks: (1) the zero-shot answer task (ZSA) as shown in Fig.~\ref{fig:zsa_example}, and (2) the zero-shot question task (ZSQ) as shown in Fig.~\ref{fig:zsq_example}.
The dataset is also helpful for detecting whether a model is biased to remember superficial relation between input and output, because such relation can not solve zero-shot transfer learning problems.
For example, when a question asks what is on the ground, the answer is likely to be snow, because this is an interesting and natural situation.

We evaluate three state-of-the-art VQA models \cite{ren2015exploring,yang2016stacked,lu2016hierarchical} on these two newly proposed tasks.
Experiments show that the testing accuracy significantly decreases on the ZSQ task.
What is even worse, the testing accuracy drops to zero on the ZSA task.
Both suggest that current models do not have the zero-shot transfer ability.
We made further analysis and found that the training data affect the biases in the networks, causing significant drop in performance.


In summary, our contributions are threefold. 
(1) We propose the problem of zero-shot transfer learning, which is an important skill for human-level Artificial Intelligence.
(2) We build the ZST-VQA dataset for this problem, and experiment on three existing methods. We show that these methods do not work well on this problem, and we analyze the reasons.
(3) This dataset is also useful to detect if a model learns to simply remember superficial input and output relations in VQA tasks.

\section{Related Work}

\textbf{Visual Question Answering.} 
VQA has improving dramatically recently~\cite{ren2015exploring,hu2017learning,johnson2017inferring}.
We briefly introduce typical VQA methods, and recommend the surveys~\cite{kafle2017visual,gupta2017survey} for a more details.
Depending on the attention usage, VQA methods can be roughly divided into three groups: (i) non-attention methods, (ii) visual attention methods and (iii) visual-text co-attention methods.
Non-attention methods include multimodal compact bilinear networks~\cite{fukui2016multimodal}, relational networks~\cite{santoro2017simple}, and Deeper LSTM Question+Image~\cite{Lu2015}. They usually produce answers through a general network architecture with attention implicitly embedded in the model.
Visual attention methods, on the contrary, utilize image-question pairs to attend to the discriminative image regions to predict the answer.
For example, stacked attention networks~\cite{yang2016stacked}, ABC-CNN~\cite{chen2015abc} and dynamic memory networks~\cite{xiong2016dynamic} all explicitly compute the visual attention by combining top-down question contexts and bottom-up image cues.
Visual-text co-attention methods such as
hierarchical co-attention networks~\cite{lu2016hierarchical}, dual attention networks~\cite{nam2016dual} and compositional attention networks~\cite{hudson2018compositional} all build explicit attention on both images and questions.
Tough results are encouraging, these methods don't address zero-shot transfer problem.


\textbf{Zero-Shot Learning.}
Zero-shot learning was early proposed by~\cite{palatucci2009zero} and soon become an interesting research problem in the cross-modal domain spanning natural language processing and computer vision~\cite{socher2013zero}, where no finite set of samples can cover the diversity of the real world and all datasets naturally follow a heavy-tail distribution with new classes appearing frequently after the training~\cite{socher2013zero,lake2018generalization}.
Usually, zero-shot learning requires to transfer knowledge from other sources such as attributes~\cite{lampert2014attribute}, word embedding~\cite{frome2013devise} or the relationship to other categories~\cite{wang2018zero} in order to predict the novel class labels.
In our dataset, novel words are embedded inside one module (i.e. questions) and we test the model's zero-shot generalization ability to the other module (i.e. answers).


\textbf{VQA datasets.}
There are many VQA datasets, such as ~\cite{johnson2017clevr,agrawal2017c} for compositionality, zero-shot VQA dataset~\cite{teney2016zero,ramakrishnan2017empirical} using extra resources, and more on the surveys~\cite{kafle2017visual,gupta2017survey}. Different from them, our dataset focus on transfer between input and output for natural human learning.

\section{Dataset Construction}

\begin{algorithm}[!t]
\caption{Steps to create the proposed ZST-VQA dataset.}
\label{algo:creation_steps}
\begin{algorithmic}[1]\small
\State Mix the original train and test samples together, and find the shared words between questions and answers.
\State Filter out stop words from the shared words list.
\State Randomly sample 10\% of words from this list with uniform distribution.
\State Equally divide the sampled words to ZSA and ZSQ words.
\State Create \textbf{ZSA test dataset} from all samples with ZSA words in answer.
\State Create \textbf{ZSQ test dataset} from all samples with ZSQ words in question.
\State Create \textbf{normal test dataset} from the original test samples that don't overlap with ZSA test and ZSQ test.
\State Create \textbf{normal train dataset} from the original train samples that don't overlap with ZSA test and ZSQ test.
\State Combine three test datasets, and label each sample with its source test dataset.
\end{algorithmic}
\end{algorithm}

We consider two scenarios:
(1) Zero-shot answer (ZSA), where a set of selected words are contained in training questions and testing answers, but not training answers. In this case, we expect a model to transfer the concept of words learned from questions to answers for correct outputs in testing. (2) Zero-shot question (ZSQ), where another set of words are contained in training answers and testing questions, but not in training questions. ZSA and ZSQ are similar tasks in the opposite direction.

\begin{table}[t]
\begin{center}
\caption{Statistics of the original VQA~\cite{antol2015vqa} and the proposed ZST-VQA dataset.}
\label{table:datasets}
\begin{tabular}{l||r|r||r|r|r|r}
\multicolumn{1}{l||}{} & \multicolumn{2}{c||}{VQA v1.0~\cite{antol2015vqa}} & \multicolumn{4}{c}{zero-shot transfer VQA} \\
\hline 
 & \multicolumn{1}{c|}{Original train} & \multicolumn{1}{c||}{Original test} & \multicolumn{1}{c|}{Normal train} & \multicolumn{1}{c|}{Normal test} & \multicolumn{1}{c|}{ZSA test} & \multicolumn{1}{c}{ZSQ test}\\
\hline
Questions & 248,349 & 121,512 & 163,281 & 80,436 & 15,302 & 34,353\\
Images & 82,783 & 40,504 & 54,427 & 26,812 & 14,549 & 29,962\\
\end{tabular}
\vspace{-0.3cm}
\end{center}
\end{table}

To do this, we first find a list of shared words between questions and answers in the original VQA dataset (including both training and testing) and filter out stop words from it.
We then uniformly sample two mutually exclusive sets of words from this list as zero-shot words for ZSA and ZSQ tasks.
Note that this is different from~\cite{teney2016zero} where only words less than 20 times are selected.
This is because of two main concerns:
(1) Zero-shot words may appear when there are not enough training data so that some words are not observed. In this case, the frequency of the zero-shot words in the test set may be small. (2) Zero-shot words may also appear when there is concept drift (i.e. new words appear over time). In this case, the frequency of zero-shot words in the test set can be high. In order to consider both cases, we hence uniformly sample zero-shot words from words shared by all questions and answers with all possible frequencies.

We then create the ZSA test set by extracting samples that contain the corresponding zero-shot words in answers. Similarly, we create the ZSQ test set with the corresponding zero-shot words in questions. The remaining training samples are used as the normal train set, and the remaining test sample as the normal test set.
To avoid the normal training set sharing images with any test set and to keep the normal test set generated in the same process as the normal training set, we discard training samples if they share the same images as ZSA or ZSQ test sets.
Algorithm~\ref{algo:creation_steps} summarize our dataset creation process and Table~\ref{table:datasets} shows the basic statistics.
Due to page limitation, please see more dataset details in the appendix.

\section{Experiments}
\subsection{Baselines}
We evaluate three typical types of VQA algorithms on our dataset: non-attention (LSTM Q+I)~\cite{Lu2015}, visual attention (SAN)~\cite{yang2016stacked}, and visual-text co-attention (HieCoAtt)~\cite{lu2016hierarchical} methods.

Deeper LSTM Question+Image (LSTM Q+I)~\cite{Lu2015} processes images and questions in different channels. The image channel extracts image feature from the last hidden layer of VGGNet~\cite{simonyan2014very}. The question channel extract question feature from the last state of a 2-layered LSTM. The image and question features are then merged by element-wise multiplication and fed into a fully-connected layer for prediction. The model is trained with cross-entropy loss.

Stacked Attention Networks (SAN)~\cite{yang2016stacked} computes a sequence of attentions for prediction. The question is converted to an embedding to obtain attention map over image feature map. The attended image and the question embedding are then combined with the encoded question as a new embedding. It repeats this process for two times, and a fully-connected layer is used for prediction.

Hierarchical Question-Image Co-attention Networks (HieCoAtt)~\cite{lu2016hierarchical} applies attentions on both images and questions. Both attentions are computed hierarchically and the features in different hierarchical levels are combined. Then it is passed to a fully-connected layer for prediction.


\begin{table}[t]
\begin{center}
\caption{Testing accuracies of baseline and extended baseline methods on zero-shot transfer VQA dataset, with reported results on VQA dataset as reference.}
\label{table:results}
\begin{tabular}{l|r|rrr|rrr}
 & & \multicolumn{3}{c|}{Baselines} & \multicolumn{3}{c}{Extended baselines} \\
\hline
 & \multicolumn{1}{c|}{VQA val} & \multicolumn{1}{c}{Normal} & \multicolumn{1}{c}{ZSA} & \multicolumn{1}{c|}{ZSQ} & \multicolumn{1}{c}{Normal} & \multicolumn{1}{c}{ZSA} & \multicolumn{1}{c}{ZSQ}\\
\hline
LSTM Q+I~\cite{Lu2015} &  54.23\% & 47.72\% & 0.00\% & 40.03\% & 46.43\% & 0.00\% & 40.14\%\\
SAN~\cite{yang2016stacked} & 55.86\% & 48.70\% & 0.00\% & 41.63\% & 46.35\% & 0.00\% & 40.54\%\\
HieCoAtt~\cite{lu2016hierarchical} & 57.09\% & 41.70\% & 0.00\% & 35.46\% & 40.10\% & 0.00\% & 33.95\%\\
\end{tabular}
\end{center}
\vspace{-0.5cm}
\end{table}

\subsection{Results}

We use publicly available original implementation for each algorithm. 
The results in Table~\ref{table:results} (middle) show 0\% accuracy in ZSA and significantly lower results than normal test set in ZSQ.
There are some obvious reasons for the performance drop.
In the baseline methods, question and answer vocabularies are constructed with training dataset. This means answer set does not contain zero-shot answers, and question vocabulary does not contain zero-shot question words. At test time, zero-shot answers will not be predicted, because the models do not have the corresponding classes, and all zero-shot question words are treated as unknown words.
For ZSA problem, even if zero-shot classes are available in training, since these methods use fully-connected layer as the last layer, the bias in this layer for the zero-shot classes will be very low, because all it always receives negative gradient. These low bias will reduce the score of the zero-shot classes during testing, and prevents zero-shot answer prediction.
Another problem is questions and answers don't share information so that a word is treated as different ones in question and answer, therefore it cannot be transferred.

To avoid these problems, we define joint vocabulary for both questions and answers, share the word embedding and the transposed weights of the last fully-connected layer, and remove bias from the last layer. The result (Table~\ref{table:results} right) shows that ZSA still has zero accuracy, and ZSQ is still significantly lower than normal test, which is consistent with previous observation.

\begin{figure}[!t]
  \centering
  \subfloat[Average scores. Implicit bias is learned and preserved to zero-shot test task.]{\includegraphics[width=0.48\textwidth]{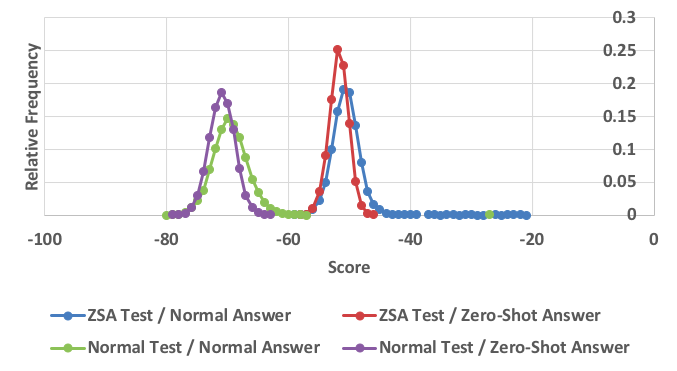}\label{fig:score_distribution}}
  \hfill
  \subfloat[Implicit bias for a sample in Fig.~\ref{fig:zsa_example}. ``sunglasses'' is predicted as ``frisbee''.]{\includegraphics[width=0.48\textwidth]{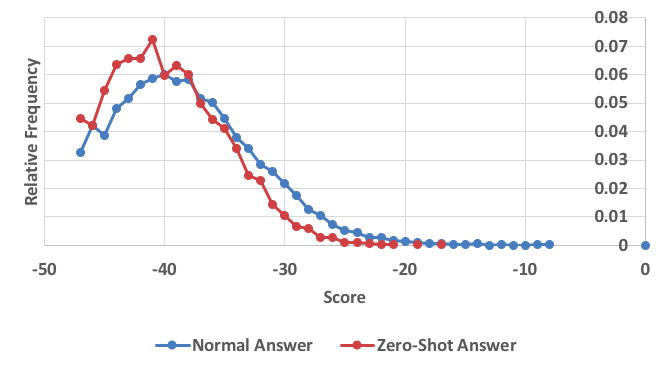}\label{fig:example_score}}
  \caption{Normalized histograms of binned scores for answers. Implicit bias assigns higher scores on normal answers (blue, green) than on zero-shot answers (red, purple).}
\vspace{-0.3cm}
\end{figure}

\section{Discussions}
The above results indicate that there should be other reasons for performance drop. 
In ZSA task, a reason might be intermediate network learns an implicit bias and always maps test input to a manifold that appeared in training. We use SAN for analysis.
In Fig.~\ref{fig:score_distribution}, we plot normalized histograms of binned average scores for answers. The scores are the last representation before the final Softmax layer, so that they are monotonic to output probabilities. We consider two test datasets for normal and ZSA tasks, and two answer sets for normal and zero-shot answers, altogether four histograms.
In normal test, the curve of normal answer set (green) is right to that of zero-shot answer set (purple). This indicates the model has learned an implicit bias to assign higher scores on normal answers. In ZSA test, the curve of normal answer set (blue) is also right to that of zero-shot answer set (red), and the relation (blue vs. red) is very similar to that in normal test (green vs. purple). This indicates that the implicit bias is preserved to ZSA task, so that zero-shot answers have low scores, and the model doesn't predict them as answers.
We also plot histograms in Fig.~\ref{fig:example_score} for a sample in Fig.~\ref{fig:zsa_example} with question [``What is on the man's face?''], and ``sunglasses'' (zero-shot answer) is predicted as ``frisbee'' (normal answer).
The plot shows the implicit bias also exists for a single sample.


In ZSQ, the performance is also worse than normal test, meaning the words aren't transferred from answers to questions very well. However, unlike ZSA task, it has much higher accuracy than zero or random prediction. This might be because questions contain many words so that zero-shot words don't influence the prediction too much.

In both ZSA and ZSQ tasks, one key problem might be compositionality, decoupling words and the way that they are processed. This is because transfer alone doesn't solve zero-shot problems, and the model needs the ability to process transferred zero-shot wards in the same manner as normal wards.
This may be achieved with some special network architectures, such as attention and memory, appropriate loss and regularization, or other types of innovations.

\section{Conclusion}
We created the ZST-VQA dataset with ZSA and ZSQ tasks.
We evaluated existing VQA algorithms on the new dataset, and found that the algorithms and their extensions do not address the zero-shot transfer problem.
We analyzed that implicit bias may be a reason of low performance, and discussed possible solutions with compositionality.
We hope this dataset will encourage the research on zero-shot transfer learning in VQA community.

\section*{Acknowledgement}
We thank Liang Zhao and Ka Yee Lun for meaningful discussions and revisions.

\clearpage

\bibliographystyle{plain}
\bibliography{nips_2018}

\begin{thebibliography}{10}

\bibitem{agrawal2017c}
Aishwarya Agrawal, Aniruddha Kembhavi, Dhruv Batra, and Devi Parikh.
\newblock C-vqa: A compositional split of the visual question answering (vqa)
  v1. 0 dataset.
\newblock {\em arXiv preprint arXiv:1704.08243}, 2017.

\bibitem{antol2015vqa}
Stanislaw Antol, Aishwarya Agrawal, Jiasen Lu, Margaret Mitchell, Dhruv Batra,
  C~Lawrence~Zitnick, and Devi Parikh.
\newblock Vqa: Visual question answering.
\newblock In {\em Proceedings of the IEEE International Conference on Computer
  Vision}, pages 2425--2433, 2015.

\bibitem{chen2015abc}
Kan Chen, Jiang Wang, Liang-Chieh Chen, Haoyuan Gao, Wei Xu, and Ram Nevatia.
\newblock Abc-cnn: An attention based convolutional neural network for visual
  question answering.
\newblock {\em arXiv preprint arXiv:1511.05960}, 2015.

\bibitem{frome2013devise}
Andrea Frome, Greg~S Corrado, Jon Shlens, Samy Bengio, Jeff Dean, Tomas
  Mikolov, et~al.
\newblock Devise: A deep visual-semantic embedding model.
\newblock In {\em Advances in neural information processing systems}, pages
  2121--2129, 2013.

\bibitem{fukui2016multimodal}
Akira Fukui, Dong~Huk Park, Daylen Yang, Anna Rohrbach, Trevor Darrell, and
  Marcus Rohrbach.
\newblock Multimodal compact bilinear pooling for visual question answering and
  visual grounding.
\newblock {\em arXiv preprint arXiv:1606.01847}, 2016.

\bibitem{gao2015you}
Haoyuan Gao, Junhua Mao, Jie Zhou, Zhiheng Huang, Lei Wang, and Wei Xu.
\newblock Are you talking to a machine? dataset and methods for multilingual
  image question.
\newblock In {\em Advances in neural information processing systems}, pages
  2296--2304, 2015.

\bibitem{gelder1990compositionality}
Tim Gelder.
\newblock Compositionality: A connectionist variation on a classical theme.
\newblock {\em Cognitive Science}, 14(3):355--384, 1990.

\bibitem{gupta2017survey}
Akshay~Kumar Gupta.
\newblock Survey of visual question answering: Datasets and techniques.
\newblock {\em arXiv preprint arXiv:1705.03865}, 2017.

\bibitem{he2016deep}
Kaiming He, Xiangyu Zhang, Shaoqing Ren, and Jian Sun.
\newblock Deep residual learning for image recognition.
\newblock In {\em Proceedings of the IEEE conference on computer vision and
  pattern recognition}, pages 770--778, 2016.

\bibitem{hu2017learning}
Ronghang Hu, Jacob Andreas, Marcus Rohrbach, Trevor Darrell, and Kate Saenko.
\newblock Learning to reason: End-to-end module networks for visual question
  answering.
\newblock {\em CoRR, abs/1704.05526}, 3, 2017.

\bibitem{hudson2018compositional}
Drew~A Hudson and Christopher~D Manning.
\newblock Compositional attention networks for machine reasoning.
\newblock {\em arXiv preprint arXiv:1803.03067}, 2018.

\bibitem{johnson2017clevr}
Justin Johnson, Bharath Hariharan, Laurens van~der Maaten, Li~Fei-Fei,
  C~Lawrence Zitnick, and Ross Girshick.
\newblock Clevr: A diagnostic dataset for compositional language and elementary
  visual reasoning.
\newblock In {\em Computer Vision and Pattern Recognition (CVPR), 2017 IEEE
  Conference on}, pages 1988--1997. IEEE, 2017.

\bibitem{johnson2017inferring}
Justin Johnson, Bharath Hariharan, Laurens van~der Maaten, Judy Hoffman,
  Li~Fei-Fei, C~Lawrence Zitnick, and Ross Girshick.
\newblock Inferring and executing programs for visual reasoning.
\newblock {\em arXiv preprint arXiv:1705.03633}, 2017.

\bibitem{kafle2017visual}
Kushal Kafle and Christopher Kanan.
\newblock Visual question answering: Datasets, algorithms, and future
  challenges.
\newblock {\em Computer Vision and Image Understanding}, 163:3--20, 2017.

\bibitem{krizhevsky2012imagenet}
Alex Krizhevsky, Ilya Sutskever, and Geoffrey~E Hinton.
\newblock Imagenet classification with deep convolutional neural networks.
\newblock In {\em Advances in neural information processing systems}, pages
  1097--1105, 2012.

\bibitem{lake2018generalization}
Brenden Lake and Marco Baroni.
\newblock Generalization without systematicity: On the compositional skills of
  sequence-to-sequence recurrent networks.
\newblock In {\em International Conference on Machine Learning}, pages
  2879--2888, 2018.

\bibitem{lampert2014attribute}
Christoph~H Lampert, Hannes Nickisch, and Stefan Harmeling.
\newblock Attribute-based classification for zero-shot visual object
  categorization.
\newblock {\em IEEE Transactions on Pattern Analysis and Machine Intelligence},
  36(3):453--465, 2014.

\bibitem{Lu2015}
Jiasen Lu, Xiao Lin, Dhruv Batra, and Devi Parikh.
\newblock Deeper lstm and normalized cnn visual question answering model.
\newblock {\em GitHub repository}, 2015.

\bibitem{lu2016hierarchical}
Jiasen Lu, Jianwei Yang, Dhruv Batra, and Devi Parikh.
\newblock Hierarchical question-image co-attention for visual question
  answering.
\newblock In {\em Advances In Neural Information Processing Systems}, pages
  289--297, 2016.

\bibitem{nam2016dual}
Hyeonseob Nam, Jung-Woo Ha, and Jeonghee Kim.
\newblock Dual attention networks for multimodal reasoning and matching.
\newblock {\em arXiv preprint arXiv:1611.00471}, 2016.

\bibitem{OpenAI_dota}
OpenAI.
\newblock Openai five.
\newblock \url{https://blog.openai.com/openai-five/}, 2018.

\bibitem{palatucci2009zero}
Mark Palatucci, Dean Pomerleau, Geoffrey~E Hinton, and Tom~M Mitchell.
\newblock Zero-shot learning with semantic output codes.
\newblock In {\em Advances in neural information processing systems}, pages
  1410--1418, 2009.

\bibitem{ramakrishnan2017empirical}
Santhosh~K Ramakrishnan, Ambar Pal, Gaurav Sharma, and Anurag Mittal.
\newblock An empirical evaluation of visual question answering for novel
  objects.
\newblock {\em arXiv preprint arXiv:1704.02516}, 2017.

\bibitem{ren2015exploring}
Mengye Ren, Ryan Kiros, and Richard Zemel.
\newblock Exploring models and data for image question answering.
\newblock In {\em Advances in neural information processing systems}, pages
  2953--2961, 2015.

\bibitem{santoro2017simple}
Adam Santoro, David Raposo, David~G Barrett, Mateusz Malinowski, Razvan
  Pascanu, Peter Battaglia, and Tim Lillicrap.
\newblock A simple neural network module for relational reasoning.
\newblock In {\em Advances in neural information processing systems}, pages
  4974--4983, 2017.

\bibitem{silver2017mastering}
David Silver, Julian Schrittwieser, Karen Simonyan, Ioannis Antonoglou, Aja
  Huang, Arthur Guez, Thomas Hubert, Lucas Baker, Matthew Lai, Adrian Bolton,
  et~al.
\newblock Mastering the game of go without human knowledge.
\newblock {\em Nature}, 550(7676):354, 2017.

\bibitem{simonyan2014very}
Karen Simonyan and Andrew Zisserman.
\newblock Very deep convolutional networks for large-scale image recognition.
\newblock {\em arXiv preprint arXiv:1409.1556}, 2014.

\bibitem{socher2013zero}
Richard Socher, Milind Ganjoo, Christopher~D Manning, and Andrew Ng.
\newblock Zero-shot learning through cross-modal transfer.
\newblock In {\em Advances in neural information processing systems}, pages
  935--943, 2013.

\bibitem{teney2016zero}
Damien Teney and Anton van~den Hengel.
\newblock Zero-shot visual question answering.
\newblock {\em arXiv preprint arXiv:1611.05546}, 2016.

\bibitem{wang2018zero}
Xiaolong Wang, Yufei Ye, and Abhinav Gupta.
\newblock Zero-shot recognition via semantic embeddings and knowledge graphs.
\newblock In {\em Proceedings of the IEEE Conference on Computer Vision and
  Pattern Recognition}, pages 6857--6866, 2018.

\bibitem{xiong2016dynamic}
Caiming Xiong, Stephen Merity, and Richard Socher.
\newblock Dynamic memory networks for visual and textual question answering.
\newblock In {\em International Conference on Machine Learning}, pages
  2397--2406, 2016.

\bibitem{yang2016stacked}
Zichao Yang, Xiaodong He, Jianfeng Gao, Li~Deng, and Alex Smola.
\newblock Stacked attention networks for image question answering.
\newblock In {\em Proceedings of the IEEE Conference on Computer Vision and
  Pattern Recognition}, pages 21--29, 2016.

\end{thebibliography}

\clearpage

\appendix

\section{Frequent Zero-Shot Words}
Zero-shot words are used to select samples for ZSA and ZSQ data sets. Some of these words often appears and some are not. To understand the effect of these words, we list frequent zero-shot words that appear in answer in ZSA data set (Table~\ref{table:frequent-zsa-words}) and in question for ZSQ data set (Table~\ref{table:frequent-zsq-words}). The lists of all zero-shot words are publicly available as a part of the dataset. 
These zero-shot words are already filtered by stop words from this address.\footnote{https://github.com/Yoast/YoastSEO.js/blob/develop/src/config/stopwords.js} 

\begin{table}
\begin{center}
\caption{Frequent zero-shot words appearing in ZSA test set answers.}
\small
\label{table:frequent-zsa-words}
\begin{tabular}{rl|rl|rl|rl|rl}
Freq. & Word & Freq. & Word & Freq. & Word & Freq. & Word & Freq. & Word\\
\hline
6,206 & 3 & 125 & closed & 74 & beige & 43 & cherry & 32 & straight\\
1,229 & 0 & 120 & sunglasses & 56 & love & 43 & bun & 32 & marble\\
1,138 & dog & 115 & plastic & 56 & center & 42 & dusk & 30 & butterfly\\
684 & stop & 115 & head & 54 & swimming & 42 & bandana & 28 & hours\\
625 & purple & 111 & chicken & 52 & sugar & 41 & serve & 28 & daisies\\
324 & male & 105 & meter & 50 & know & 41 & east & 28 & bamboo\\
302 & bat & 102 & tomato & 49 & parrot & 39 & 35 & 27 & remotes\\
241 & picture & 89 & overcast & 45 & ford & 38 & surfer & 27 & cheesecake\\
137 & tower & 85 & blanket & 44 & p & 34 & stand & 26 & natural\\
126 & giraffes & 76 & tea & 43 & tulips & 32 & tape & 24 & tour\\
\end{tabular}
\end{center}
\end{table}

\begin{table}
\begin{center}
\caption{Frequent zero-shot words appearing in ZSQ test set questions.}
\small
\label{table:frequent-zsq-words}
\begin{tabular}{rl|rl|rl|rl|rl}
Freq. & Word & Freq. & Word & Freq. & Word & Freq. & Word & Freq. & Word\\
\hline
3,205 & sport & 464 & use & 296 & carrying & 155 & crossing & 92 & popular\\
1,153 & appear & 461 & chair & 287 & face & 150 & tires & 92 & planning\\
976 & fire & 430 & parked & 283 & writing & 148 & cabinets & 88 & batters\\
898 & pattern & 415 & coming & 230 & television & 145 & beds & 85 & enjoy\\
845 & material & 399 & buses & 223 & vegetarian & 127 & mountain & 85 & dirt\\
689 & bed & 395 & take & 220 & beside & 119 & levels & 85 & carpet\\
669 & facing & 357 & utensil & 201 & graffiti & 117 & catcher & 83 & slice\\
612 & big & 335 & dish & 167 & foot & 113 & falling & 80 & salad\\
565 & trying & 321 & sink & 163 & couch & 101 & faces & 79 & square\\
474 & sandwich & 317 & three & 160 & silver & 98 & fireplace & 78 & roll\\
\end{tabular}
\end{center}
\end{table}

\section{Data Movement from Original to New Datasets}
To better understand the connection between original VQA dataset and the new dataset, we summarized the movement of samples between the two sets(Table~\ref{table:data_movement}). It shows that for both questions and images, there is no intersection between original train and normal test dataset, or between original test and normal train dataset. Also, ZSA and ZSQ datasets have the same ratio of the samples from original train and test sets.

\begin{table}[!t]
\begin{center}
\caption{Movement of question samples from original train and test data sets to new datasets.
}
\label{table:data_movement}
\begin{tabular}{l|rr|rr}
 & \multicolumn{2}{c|}{Questions} & \multicolumn{2}{c}{Images} \\
 & Train & Test & Train & Test\\
\hline
Normal Train & 163,281 & 0 & 54,427 & 0\\
Normal Test  & 0 & 80,436 & 0 & 26,812\\
ZSA Test & 10,360 & 4,942 & 9,860 & 4,689\\
ZSQ Test & 23,145 & 11,208 & 20,165 & 9,797\\
\end{tabular}
\end{center}
\end{table}

\end{document}